\documentclass[10pt,twocolumn,letterpaper]{article}

\usepackage{iccv}
\usepackage{times}
\usepackage{epsfig}
\usepackage{graphicx}
\usepackage{amsmath}
\usepackage{amssymb}
\usepackage{multirow}
\usepackage{booktabs}

\usepackage[pagebackref=true,breaklinks=true,letterpaper=true,colorlinks,bookmarks=false]{hyperref}

\iccvfinalcopy 

\def\iccvPaperID{6414} 

\newcommand{\calL}{\ensuremath{{\cal L}}}
\newcommand{\calX}{\ensuremath{{\cal X}}}

\ificcvfinal\fi

\begin{document}

\title{PAMTRI: Pose-Aware Multi-Task Learning for Vehicle Re-Identification \\Using Highly Randomized Synthetic Data}

\author{Zheng Tang\thanks{Work done as an intern at NVIDIA. Zheng is now with Amazon.} \quad Milind Naphade \quad Stan Birchfield  \quad Jonathan Tremblay \\
William Hodge \quad Ratnesh Kumar \quad Shuo Wang \quad Xiaodong Yang \\
NVIDIA
}

\maketitle
\ificcvfinal\thispagestyle{empty}\fi

\begin{abstract}
   In comparison with person re-identification (ReID), which has been widely studied in the research community, vehicle ReID has received less attention. 
   Vehicle ReID is challenging due to 1) high intra-class variability (caused by the dependency of shape and appearance on viewpoint), and 2) small inter-class variability (caused by the similarity in shape and appearance between vehicles produced by different manufacturers). 
   To address these challenges, we propose a Pose-Aware Multi-Task Re-Identification (PAMTRI) framework. 
   This approach includes two innovations compared with previous methods.
   First, it overcomes viewpoint-dependency by explicitly reasoning about vehicle pose and shape via keypoints, heatmaps and segments from pose estimation. 
   Second, it jointly classifies semantic vehicle attributes (colors and types) while performing ReID, through multi-task learning with the embedded pose representations. 
   Since manually labeling images with detailed pose and attribute information is prohibitive, we create a large-scale highly randomized synthetic dataset with automatically annotated vehicle attributes for training. 
   Extensive experiments validate the effectiveness of each proposed component, showing that PAMTRI achieves significant improvement over state-of-the-art on two mainstream vehicle ReID benchmarks: VeRi and CityFlow-ReID. 
   Code and models are available at \href{https://github.com/NVlabs/PAMTRI}{https://github.com/NVlabs/PAMTRI}.
   
\end{abstract}
\section{Introduction}

\begin{figure}[t]
\begin{center}
\includegraphics[width=0.95\linewidth]{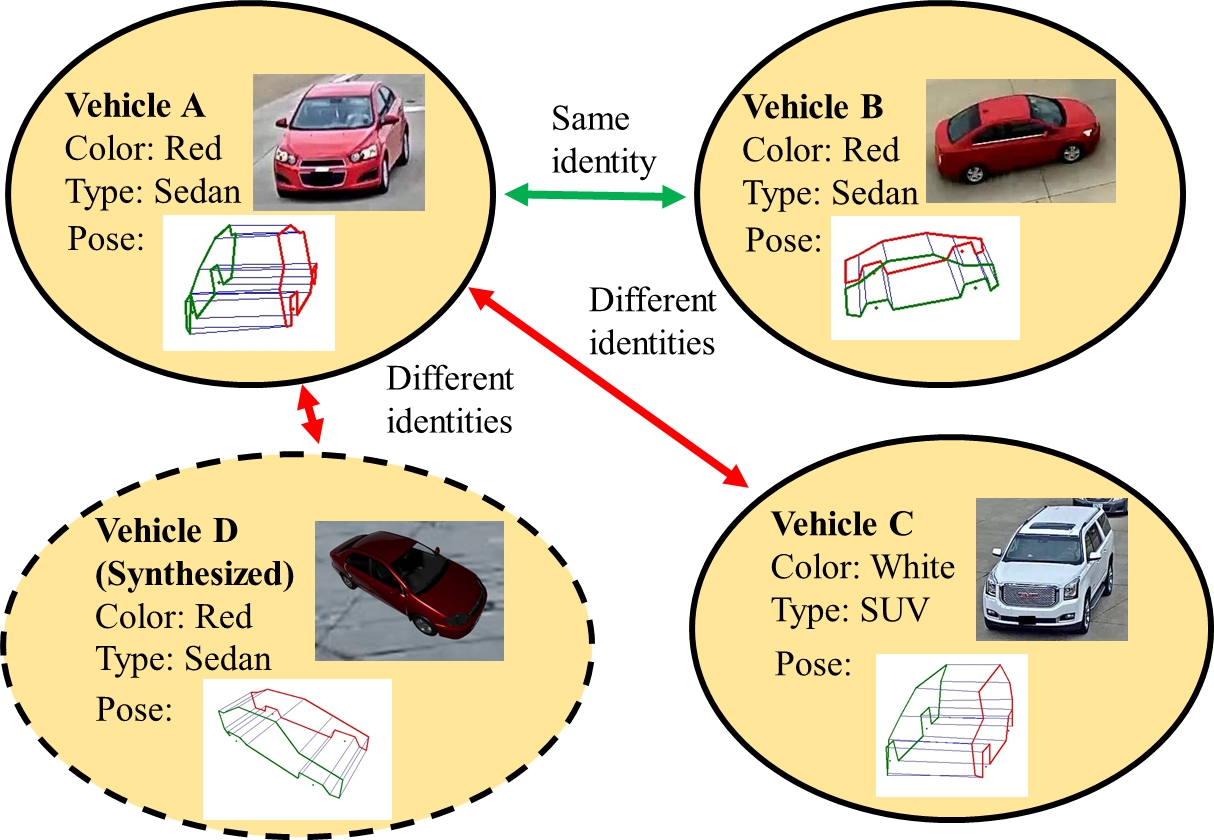}
\end{center}
   \caption{The problem of vehicle ReID involves identifying the same vehicle across different viewing perspectives and cameras, based solely on appearance in the images.  Our approach uses multi-task learning to leverage information about the vehicle's pose and semantic attributes (color and type). Synthetic data play a key role in training, enabling highly detailed annotations to be generated automatically and inexpensively. \textbf{Best viewed in color.}}
\label{fig:illustration}
\end{figure}

The wide deployment of traffic cameras presents an immense opportunity for video analytics in a variety of applications such as logistics, transportation, and smart cities.  
A particularly crucial problem in such analytics is the cross-camera association of targets like pedestrians and vehicles, \textit{i.e.}, re-identification (ReID), which is illustrated in Fig.~\ref{fig:illustration}. 

Although both pedestrians and vehicles are  common objects in smart city applications, in recent years most attention has been paid to person ReID. 
This is mainly due to the abundance of well-annotated pedestrian data, along with the historical focus of computer vision research on human faces and bodies. 
Furthermore, compared to pedestrians, vehicle ReID is arguably more challenging due to high \emph{intra-class variability} caused by the variety of shapes from different viewing angles, coupled with small \emph{inter-class variability} since car models produced by various manufacturers are limited in their shapes and colors.
To verify this intuition, we compared the feature distributions in both person-based and vehicle-based ReID tasks.
Specifically, we used GoogLeNet~\cite{Szegedy15Going} pre-trained on ImageNet~\cite{Deng09ImageNet} to extract 1,024-dimensional features from Market1501~\cite{Zheng15Scalable} and CityFlow-ReID~\cite{Tang19CityFlow}, respectively. 
For each dataset, the ratio of the intra- to inter-class variability (based on Euclidean feature distance) was calculated. 
The results are as follows: 0.921 for pedestrians (Market1501) and 0.946 for vehicles (CityFlow-ReID), which support the notion that vehicle-based ReID is more difficult.
Although license plates could potentially be useful to identify each vehicle, they often cannot be read from traffic cameras due to occlusion, oblique viewpoint, or low image resolution, and they present privacy concerns. 

Recent methods to vehicle ReID exploit feature learning~\cite{Wang17Orientation, Zhou18Aware, Zhou18Vehicle} and/or distance metric learning~\cite{Bai18Group, Kumar19Vehicle,Liu16Deep} to train deep neural networks (DNNs) to distinguish between vehicle pairs, but the current state-of-the-art performance is still far from its counterpart in person ReID~\cite{Zheng19Joint}. 
Moreover, it has been shown~\cite{Tang19CityFlow} that directly using state-of-the-art person ReID methods for vehicles does not close this gap, indicating fundamental differences between the two tasks. 
We believe the key to vehicle ReID is to exploit viewpoint-invariant information such as color, type, and deformable shape models encoding pose.  
To jointly learn these attributes along with pose information, we propose to use synthetic data to overcome the prohibitive cost of manually labeling real training images with such detailed information. 

In this work, we propose a novel framework named PAMTRI, for Pose-Aware Multi-Task Re-Identification. 
Our major contribution is threefold: 

\begin{enumerate}
  \item PAMTRI embeds keypoints, heatmaps and segments from pose estimation into the multi-task learning pipeline for vehicle ReID, which guides the network to pay attention to viewpoint-related information. 
  
  \item PAMTRI is trained with large-scale synthetic data that include randomized vehicle models, color and orientation under different backgrounds, lighting conditions and occlusion. Annotations of vehicle identity, color, type and 2D pose are automatically generated for training.
  
  \item Our proposed method achieves significant improvement over the state-of-the-art on two mainstream benchmarks: VeRi~\cite{Liu16A} and CityFlow-ReID~\cite{Tang19CityFlow}. Additional experiments validate that our unique architecture exploiting explicit pose information, along with our use of randomized synthetic data for training, are key to the method's success.
\end{enumerate}
\section{Related work}

\textbf{Vehicle ReID.}  
Among the earliest attempts for vehicle ReID that involve deep learning, Liu \etal~\cite{Liu16A, Liu17PROVID} propose a progressive framework that employs a Siamese neural network with contrastive loss for training, and they also introduced VeRi~\cite{Liu16A} as the first large-scale benchmark specifically for vehicle ReID. 
Bai \etal~\cite{Bai18Group} and Kumar \etal~\cite{Kumar19Vehicle} also take advantage of distance metric learning by extending the success of triplet embedding in person ReID~\cite{Hermans17In} to the vehicle-based task. 
Especially, the batch-sampling variant from Kumar \etal is the current state-of-the-art on both VeRi and CityFlow-ReID~\cite{Tang19CityFlow}, the latter being a subset of a recent multi-target multi-camera vehicle tracking benchmark. 
On the other hand, some methods focus on exploiting viewpoint-invariant features, \eg, the approach by Wang \etal~\cite{Wang17Orientation} that embeds local region features from extracted vehicle keypoints for training with cross-entropy loss. 
Similarly, Zhou \etal~\cite{Zhou18Aware, Zhou18Vehicle} use a generative adversarial network (GAN) to generate multi-view features to be selected by a viewpoint-aware attention model, in which attribute classification is also trained through the discriminative net. 
In addition, Yan \etal~\cite{Yan17Exploiting} apply multi-task learning to address multi-grain ranking and attribute classification simultaneously, but the search for visually similar vehicles is different from our goal of ReID. 
To our knowledge, none of the methods jointly embody pose information and multi-task learning to address vehicle ReID.  

\textbf{Vehicle pose estimation.}  
Vehicle pose estimation via deformable (\ie, keypoint-based) modeling is a promising approach to deal with viewpoint information.
In \cite{Tang17MultipleKernel}, Tang \etal propose to use a 16-keypoint-based car model generated from evolutionary optimization to build multiple kernels for 3D tracking.
Ansari \etal~\cite{Ansari18The} designed a more complex vehicle model with 36 keypoints for 3D localization and shape estimation from a dash camera. 
The ReID method by Wang \etal~\cite{Wang17Orientation} also employs a 20-keypoint model to extract orientation-based features for region proposal.
However, instead of explicitly locating keypoint coordinates, their network is trained for estimating response maps only, and semantic attributes are not exploited in their framework. 
Other methods can directly regress to the car pose with  6 degrees of freedom (DoF) \cite{Kundu183DRCNN,Manhardt19ROI10D,Mousavian173D, Wohlhart15Learning}, but they are limited for our purposes as detailed vehicle shape modeling via keypoints is not provided. 

\begin{figure*}[t]
\begin{center}
\includegraphics[width=0.8\linewidth]{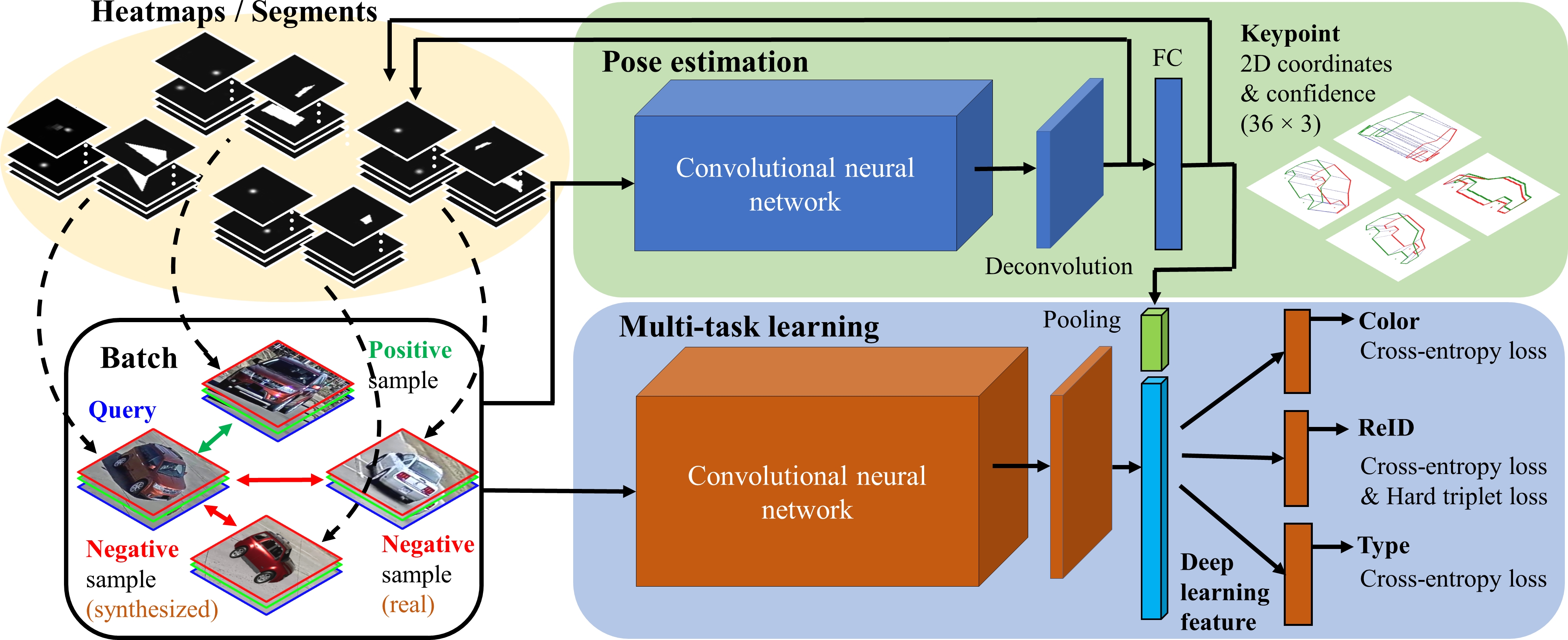}
\end{center}
   \caption{Overview of the proposed method. Each training batch includes both real and synthesized images. To embed pose information for multi-task learning, the heatmaps or segments output by a pre-trained network are stacked with the original RGB channels as input. The estimated keypoint coordinates and confidence scores are also concatenated with deep learning features for ReID and attribute (color and type) classification. The pose estimation network (top, blue) is based on HRNet~\cite{Sun19Deep}, while the multi-task learning network (bottom, orange) is based on DenseNet121~\cite{Huang17Densely}.  \textbf{Best viewed in color.}}
\label{fig:flowdiagram}
\end{figure*}

\textbf{Synthetic data.}  
To generate sufficiently detailed labels on training images, our approach leverages synthetic data.  
Our method is trained on a mix of rendered and real images. 
This places our work in the context of other research on using simulated data to train DNNs.  
A popular approach to overcome the so-called \emph{reality gap} is domain randomization  \cite{Tobin17Domain, Tremblay18Training}, in which a model is trained with extreme visual variety so that when presented with a real-world image the model treats it as just another variation. 
Synthetic data have been successfully applied to a variety of problems, such as optical flow \cite{Mayer16A}, car detection \cite{Prakash19Structured}, object pose estimation \cite{Sundermeyer18Implicit,Tremblay18Deep}, vision-based robotic manipulation  \cite{James17Transferring,Tobin17Domain}, and robotic control \cite{Chebotar18Closing,Tan18SimToReal}. 
We extend this research to ReID and semantic attribute understanding.

\section{Methodology}

In this section, we describe the algorithmic design of our proposed PAMTRI framework. An overview flow diagram of the proposed system is presented in Fig.~\ref{fig:flowdiagram}.

\subsection{Randomized synthetic dataset}

Besides vehicle identities, our approach requires additional labels of vehicle attributes and locations of keypoints.
These values, particularly the keypoints, would require considerable, even prohibitive effort, if annotated manually.  
To overcome this problem, we generated a large-scale synthetic dataset by employing 
our deep learning dataset synthesizer (NDDS)~\cite{To18NDDS} to create a randomized environment in Unreal Engine 4 (UE4), into which 3D vehicle meshes from~\cite{Prakash19Structured} were imported. 
We added to NDDS the ability to label and export specific 3D locations, \ie, keypoints (denoted as \emph{sockets} in UE4), on a CAD model.
As such we manually annotated each vehicle model with the 36 3D keypoints defined by Ansari \etal~\cite{Ansari18The}; the projected 2D locations were then output with the synthesized images. 
For randomization we used 42 vehicle 3D CAD models with 10 body colors.  
To train the data for ReID, we define a unique identity for each combination of vehicle model with a particular color.
The final generated dataset consists of 41,000 unique images with 402 identities,\footnote{The concrete mixer truck and the school bus did not get color variation and as such we exported 500 unique images for each of them. 100 images were generated for each of the remaining identities.} including the following annotations: 
keypoints, orientation, and vehicle attributes (color and type). 
When generating the dataset, background images were sampled from CityFlow~\cite{Tang19CityFlow}, and 
we also randomized the vehicle position and intensity of light. 
Additionally, during training we perform randomized post-processing such as scaling, cropping, horizontal flipping, and adding occlusion.
Some examples are shown in Fig.~\ref{fig:vcityflow}. 

\subsection{Vehicle pose estimation}

To leverage viewpoint-aware information for multi-task learning, 
we train a robust DNN for extracting pose-related representations. 
Similar to Tremblay \etal \cite{Tremblay18Training} we mix real and synthetic data to bridge the reality gap. 
More specifically, in each dataset, we utilize the pre-trained model~\cite{Ansari18The} to process sampled images and manually select about 10,000 successful samples as real training data. 

Instead of using the stacked hourglass network~\cite{Newell16Stacked} as the backbone like previous approaches~\cite{Ansari18The,Wang17Orientation}, we modify the state-of-the-art DNN for human pose estimation, HRNet~\cite{Sun19Deep}, for our purposes. 
Compared to the stacked hourglass architecture and other methods that recover high-resolution representations from low-resolution representations, HRNet maintains high-resolution representations and gradually add high-to-low resolution sub-networks with multi-scale fusion.
As a result, the predicted keypoints and heatmaps are more accurate and spatially more precise, which benefit our embedding for multi-task learning. 

\begin{figure}[t]
\begin{center}
\includegraphics[width=1.0\linewidth]{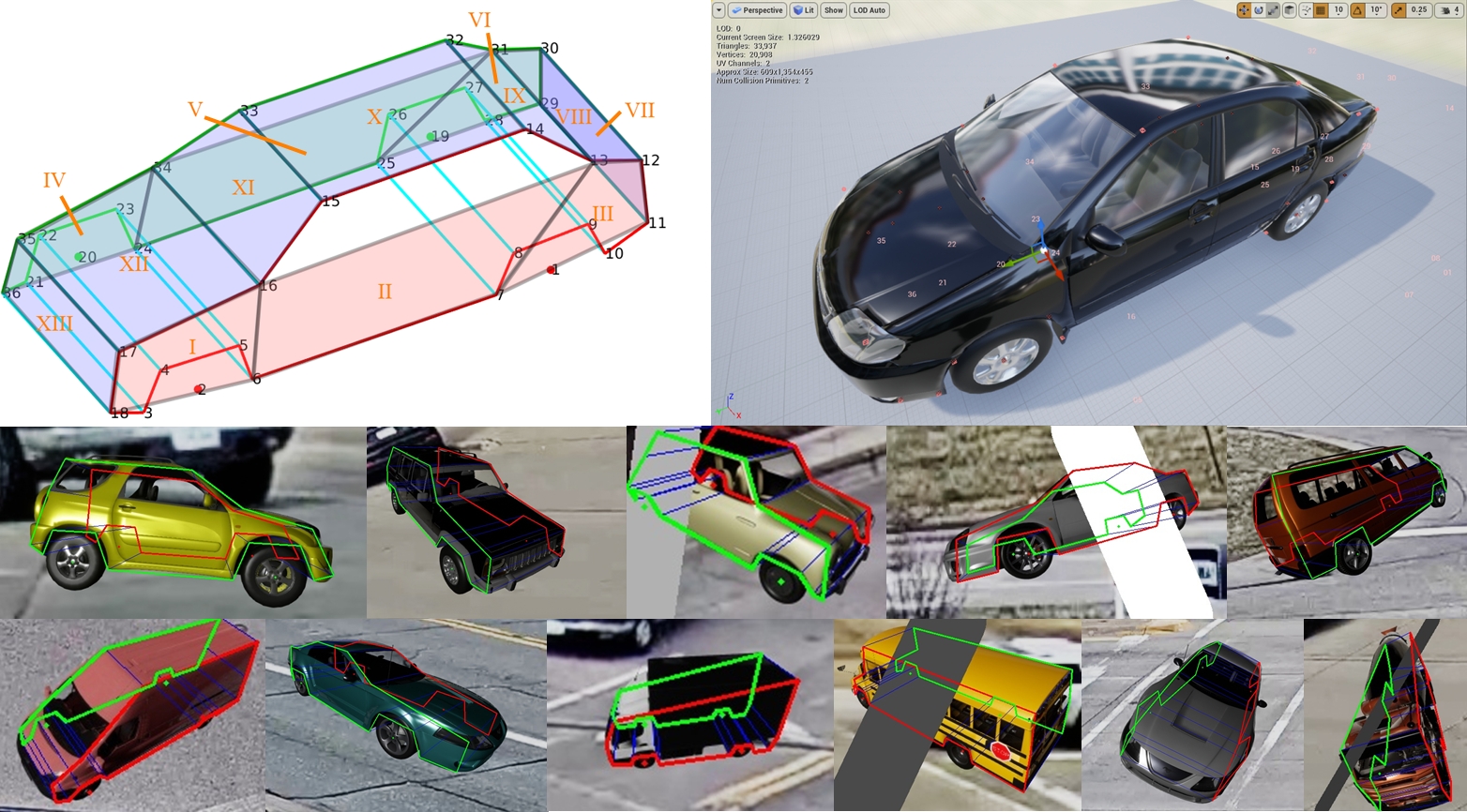}
\end{center}
   \caption{{\sc Top-left:}  The 36-keypoint model from Ansari \etal~\cite{Ansari18The} with 13 segments defined by us. {\sc Top-right:}  3D keypoints selected in UE4. {\sc Bottom:}  Example images from our randomized synthetic dataset for training, with automatically annotated poses overlaid. \textbf{Best viewed in color.}}
\label{fig:vcityflow}
\end{figure}

We propose two ways to embed the vehicle pose information as additional channels at the input layer of the multi-task network, based on heatmaps and segments, respectively. 
In one approach, after the final deconvolutional layer, we extract the 36 heatmaps for each of the keypoints used to capture the vehicle shape and pose. 
In the other approach, the predicted keypoint coordinates from the final fully-connected (FC) layer are used to segment the vehicle body.  For example, in Fig.~\ref{fig:vcityflow} the keypoints \#16, \#17, \#35 and \#34 from the deformable model form a segment that represents the car hood. 
Accordingly, we define 13 segmentation masks for each vehicle (see Fig.~\ref{fig:vcityflow} {\sc Top-left}), where those formed by keypoint(s) with low confidence are set to be blank.  
The feedback of heatmaps or segments from the pose estimation network is then scaled and appended to the original RGB channels for further processing. 
We also send the explicit keypoint coordinates and confidence to the multi-task network for further embedding.

\subsection{Multi-task learning for vehicle ReID} \label{sec:mtlearning}

Pose-aware representations are beneficial to both ReID and attribute classification tasks. 
First, vehicle pose describes the 3D shape model that is invariant to the camera viewpoint, and thus the ReID sub-branch can learn to relate features from different views. 
Second, the vehicle shape is directly connected with the car type to which the target belongs. 
Third, the segments by 2D keypoints enable the color classification sub-branch to extract the main vehicle color while neglecting the non-painted areas such as windshields and wheels. 

Hence, we embed the predicted keypoints and heatmaps (or segments) into our multi-task network to help guide the attention to viewpoint-related representations. 
First, all the feedback heatmaps/segments from pose estimation are stacked with the RGB channels of the original input to form a new image. 
Accordingly, we modified the first layer of our backbone convolutional neural network (CNN) based on DenseNet121~\cite{Huang17Densely} to allow additional input channels. 
While we use the pre-trained weights for the RGB channels, the new channels are initialized with Gaussian random weights. 
The stacked image provides the DNN with extra information about vehicle shape, and thus helps the feature extraction to concentrate on viewpoint-aware representations. 
Both synthesized and real identities are batched together and sent to the backbone CNN. 

To the deep learning feature vector extracted from the final pooling layer, we append the keypoint coordinates and confidence scores from pose prediction, which are normalized between -0.5 and 0.5. 
Since the keypoints are explicitly represented and ordered, they enable the neural network to learn a more reliable shape description in the final FC layers for multi-task learning. 
Finally, the concatenated feature vector is fed to three separate branches for multi-task learning, including a branch for vehicle ReID and two other branches for color and type classification. 

The final loss function of our network is the combined loss of the three tasks. 
For vehicle ReID, the hard-mining triplet loss is combined with cross-entropy loss to jointly exploit distance metric learning and identity classification, described as follows:
\begin{equation} \label{eq:lossid}
\calL_{\text{ID}} = \lambda_{\text{htri}} \calL_{\text{htri}} \left( a, p, n \right) + \lambda_{\text{xent}} \calL_{\text{xent}} \left( y, \hat{y} \right),
\end{equation}
where $\calL_{\text{htri}} \left( a, p, n \right)$ is the hard triplet loss with $a$, $p$ and $n$ as anchor, positive and negative samples, respectively:
\begin{equation}
\calL_{\text{htri}} \left( a, p, n \right) = \left[ \alpha + \max{\left(D_{ap}\right)} - \min{\left(D_{an}\right)} \right]_{+},
\end{equation}
in which $\alpha$ is the distance margin, $D_{ap}$ and $D_{an}$ are the distance metrics between the anchor and all positive/negative samples in feature space, and $[\cdot]_+$ indicates $\max(\cdot,0)$; and $\calL_{\text{xent}} \left( y, \hat{y} \right)$ is the cross entropy loss:
\begin{equation}
\calL_{\text{xent}} \left( y, \hat{y} \right) = -\frac{1}{N} \sum_{i=1}^{N} y_{i} \log{\left(\hat{y}_{i}\right)},
\end{equation}
where $y$ is the ground-truth vector, $\hat{y}$ is the estimation, and $N$ is the number of classes (in our case IDs). In Eq.~\eqref{eq:lossid}, $\lambda_{\text{htri}}$ and $\lambda_{\text{xent}}$ are the regularization factors, both set to 1. 

For the other two sub-tasks of attribute classification, we again employ the cross-entropy loss as follows:
\begin{align}
\calL_{\text{color}} &= \calL_{\text{xent}} \left( y_{\text{color}}, \hat{y}_{\text{color}} \right), \\
\calL_{\text{type}} &= \calL_{\text{xent}} \left( y_{\text{type}}, \hat{y}_{\text{type}} \right).
\end{align}

The final loss is the weighted combination of all tasks: 
\begin{equation}
\calL\left( \theta, \calX \right) = \calL_{\text{ID}} + \lambda_{\text{color}} \calL_{\text{color}} + \lambda_{\text{type}} \calL_{\text{type}},
\end{equation}
where $\calX = \{ \left( x_{i}, y_{i} \right) \}$ represents the input training set and $\theta$ is the set of network parameters. 
Following the practice of other researchers~\cite{Lin17Improving, Sener18MultiTask}, we set the regularization parameters of both $\lambda_{\text{color}}$ and $\lambda_{\text{type}}$ to be much lower than 1, in our case 0.125. 
This is because, in some circumstances, vehicle ReID and attribute classification are conflicting tasks, \textit{i.e.}, two vehicles of the same color and/or type may not share the same identity. 

At the testing stage, the final ReID classification layer is removed. 
For each vehicle image a 1024-dimensional feature vector is extracted from the last FC layer.
The features from each pair of query and test images are compared using Euclidean distance to determine their similarity. 
\section{Evaluation}

In this section, we present the datasets used for evaluating our proposed approach, the implementation details, experimental results showing state-of-the-art performance, and a detailed analysis of the effects of various components of PAMTRI. 

\begin{table}
\begin{footnotesize}
\begin{center}
\begin{tabular}{lccccc}
\toprule
Dataset & \# total & \# train & \# test & \# query & \# total\\
 & IDs & IDs & IDs & images & images\\
\midrule
VeRi & 776 & 576 & 200 & 1,678 & 51,038 \\
CityFlow-ReID & 666 & 333 & 333 & 1,052 & 56,277 \\
Synthetic & 402 & 402 & -- & -- & 41,000 \\
\bottomrule
\end{tabular}
\end{center}
\caption{Statistics of the datasets used for training and evaluation.}
\label{tab:dataset}
\end{footnotesize}
\end{table}

\subsection{Datasets and evaluation protocol}

Our PAMTRI system was evaluated on two mainstream large-scale vehicle ReID benchmarks, namely, VeRi~\cite{Liu16A} and CityFlow-ReID~\cite{Tang19CityFlow}, whose statistics are summarized in Tab.~\ref{tab:dataset} together with the details of the synthetic data we generated for training. 

VeRi~\cite{Liu16A} has been widely used by most recent research in vehicle ReID, as it provides multiple views of vehicles captured from 20 cameras. 
CityFlow-ReID~\cite{Tang19CityFlow} is a subset of the recent multi-target multi-camera vehicle tracking benchmark, CityFlow, which has been adopted for the AI City Challenge~\cite{Naphade19The} at CVPR 2019.  
The latter is significantly more challenging, as the footage is captured with more cameras (40) in more diverse environments (residential areas and highways). 
Unlike VeRi, the original videos are available in CityFlow, which enable us to extract background images for randomization to generate realistic synthetic data. 
Whereas the color and type information is available with the VeRi dataset, such attribute annotation is not provided by CityFlow.
Hence, another contribution of this work is that we manually labeled vehicle attributes (color and type) for all the 666 identities in CityFlow-ReID. 

In our experiments, we strictly follow the evaluation protocol proposed in Market1501~\cite{Zheng15Scalable} measuring the mean Average Precision (mAP) and the rank-$K$ hit rates. 
For mAP, we compute the mean of all queries' average precision, \textit{i.e.}, the area under the Precision-Recall curve.
The rank-$K$ hit rate denotes the possibility that at least one true positive is ranked within the top $K$ positions. 
When all the rank-$K$ hit rates are plotted against $K$, we have the Cumulative Matching Characteristic (CMC). 
In addition, rank-$K$ mAP is introduced in~\cite{Tang19CityFlow} that measures the mean of average precision for each query considering only the top $K$ matches. 

\subsection{Implementation details}

\begin{table}
\begin{footnotesize}
\begin{center}
\begin{tabular}{lcccc}
\toprule
Method & mAP & Rank-1 & Rank-5 & Rank-20 \\
\midrule
FACT~\cite{Liu16Deep} & 18.73 & 51.85 & 67.16 & 79.56 \\
PROVID*~\cite{Liu17PROVID} & 48.47 & 76.76 & 91.40 & - \\
OIFE~\cite{Wang17Orientation} & 48.00 & 65.92 & 87.66 & 96.63 \\
PathLSTM*~\cite{Shen17Learning} & 58.27 & 83.49 & 90.04 & 97.16 \\
GSTE~\cite{Bai18Group} & 59.47 & \textbf{96.24} & \textbf{98.97} & - \\
VAMI~\cite{Zhou18Aware} & 50.13 & 77.03 & 90.82 & 97.16 \\
VAMI*~\cite{Zhou18Aware} & 61.32 & 85.92 & 91.84 & 97.70 \\
ABLN~\cite{Zhou18Vehicle} & 24.92 & 60.49 & 77.33 & 88.27 \\
BA~\cite{Kumar19Vehicle} & 66.91 & 90.11 & 96.01 & 98.27 \\
BS~\cite{Kumar19Vehicle} & 67.55 & 90.23 & 96.42 & 98.63 \\
\midrule
RS & 63.76 & 90.70 & 94.40 & 97.47 \\
RS+MT & 66.18 & 91.90 & 96.90 & 98.99 \\
RS+MT+K & 68.64 & 91.60 & 96.78 & 98.75 \\
RS+MT+K+H & 71.16 & 92.74 & 96.68 & 98.40 \\
RS+MT+K+S & \textbf{71.88} & 92.86 & 96.97 & 98.23 \\
\midrule
RS w/ Xent only & 56.52 & 83.41 & 92.07 & 97.02 \\
RS w/ Htri only & 47.50 & 73.54 & 87.25 & 96.01 \\
RS+MT w/ DN201 & 64.42 & 90.58 & 96.36 & 98.81 \\
R+MT+K & 65.44 & 90.94 & 96.72 & \textbf{99.11} \\
\bottomrule
\end{tabular}
\end{center}
\caption{Experimental comparison of state-of-the-art in vehicle ReID on VeRi~\cite{Liu16Deep}. All values are shown as percentages. For our proposed method, MT, K, H, S, RS and R respectively denote multi-task learning, explicit keypoints embedded, heatmaps embedded, segments embedded, training with both real and synthetic data, and training with real data only. Xent, Htri and DN201 stand for cross-entropy loss, hard triplet loss and DenseNet201, respectively. (*) indicates the usage of spatio-temporal information.}
\label{tab:veri}
\end{footnotesize}
\end{table}

\textbf{Training for multi-task learning.} 
Leveraging the off-the-shelf implementation in~\cite{Zhou18Torchreid}, we use DenseNet121~\cite{Huang17Densely} as our backbone CNN for multi-task learning, whose initial weights are from the model pre-trained on ImageNet~\cite{Deng09ImageNet}. 
The input images are resized to $256 \times 256$ and the training batch size is set as 32. 
We utilize the Adam optimizer~\cite{Kingma14Adam} to train the base model for 60 maximum epochs. 
The initial learning rate was set to 3e-4, which decays to 3e-5 and 3e-6 at the 20$^\text{th}$ and 40$^\text{th}$ epochs, respectively. 
For multi-task learning the dimension of the last FC layer for ReID is 1,024, whereas the two FC layers for attribute classification share the size of 512 each. 
For all the final FC layers, we adopt the leaky rectified linear unit (Leaky ReLU)~\cite{Xu15Empirical} as the activation function. 

\textbf{Training for pose estimation.} 
The state-of-the-art HRNet~\cite{Sun19Deep} for human pose estimation is used as our backbone for vehicle pose estimation, which is built upon the original implementation by Sun \etal. 
Again we adopt the pre-trained weights on ImageNet~\cite{Deng09ImageNet} for initialization.
Each input image is also resized to $256 \times 256$ and the size of the heatmap/segment output is $64 \times 64$. 
We set the training batch size to be 32, and the maximum number of epochs is 210 with learning rate of 1e-3. 
The final FC layer is adjusted to output a $108$-dimensional vector, as our vehicle model consists of 36 keypoints in 2D whose visibility (indicated by confidence scores) is also computed. 

\begin{table}
\begin{footnotesize}
\begin{center}
\begin{tabular}{lcccc}
\toprule
Method & mAP (\emph{r100}) & Rank-1 & Rank-5 & Rank-20 \\
\midrule
FVS~\cite{Tang18SingleCamera} & 6.33 (5.08) & 20.82 & 24.52 & 31.27 \\
Xent~\cite{Zhou18Torchreid} & 23.18 (18.62) & 39.92 & 52.66 & 66.06 \\
Htri~\cite{Zhou18Torchreid} & 30.46 (24.04) & 45.75 & 61.24 & 75.94 \\
Cent~\cite{Zhou18Torchreid} & 10.73 (9.49) & 27.92 & 39.77 & 52.83 \\
Xent+Htri~\cite{Zhou18Torchreid} & 31.02 (25.06) & 51.69 & 62.84 & 74.91 \\
BA~\cite{Kumar19Vehicle} & 31.30 (25.61) & 49.62 & 65.02 & 80.04 \\
BS~\cite{Kumar19Vehicle} & 31.34 (25.57) & 49.05 & 63.12 & 78.80 \\
\midrule
RS & 31.41 (25.66) & 50.37 & 61.48 & 74.26 \\
RS+MT & 32.80 (27.09) & 50.93 & 66.09 & 79.46 \\
RS+MT+K & 37.18 (31.03) & 55.80 & 67.49 & \textbf{81.08} \\
RS+MT+K+H & \textbf{40.39} (\textbf{33.81}) & \textbf{59.70} & \textbf{70.91} & 80.13 \\
RS+MT+K+S & 38.64 (32.67) & 57.32 & 68.44 & 79.37 \\
\midrule
RS w/ Xent only & 29.59 (23.74) & 41.91 & 56.77 & 73.95 \\
RS w/ Htri only & 28.09 (21.95) & 40.02 & 56.94 & 74.05 \\
RS+MT w/ DN201 & 33.18 (27.10) & 51.80 & 65.49 & 79.08 \\
R+MT+K & 36.67 (30.57) & 54.56 & 66.54 & 80.89 \\
\bottomrule
\end{tabular}
\end{center}
\caption{Experimental comparison of state-of-the-art in vehicle ReID on CityFlow-ReID~\cite{Tang19CityFlow}. All values are shown as percentages, with \emph{r100} indicating the rank-100 mAP. For our proposed method, MT, K, H, S, RS and R respectively denote multi-task learning, explicit keypoints embedded, heatmaps embedded, segments embedded, training with both real and synthetic data, and training with real data only. Xent, Htri and DN201 stand for cross-entropy loss, hard triplet loss and DenseNet201, respectively.}
\label{tab:cityflow}
\end{footnotesize}
\end{table}

\subsection{Comparison of ReID with the state-of-the-art}

Tab.~\ref{tab:veri} compares PAMTRI's performance with state-of-the-art in vehicle ReID. 
Notice that our method outperforms all the others in terms of the mAP metric. 
Although GSTE~\cite{Bai18Group} achieves higher rank-$K$ hit rates, its mAP score is lower than ours by about 10\%, which demonstrates our robust performance at all ranks. 
Note also that GSTE exploits additional group information, \textit{i.e.}, signatures of the same identity from the same camera are grouped together, which is not required in our proposed scheme. 
Moreover, VeRi also provides spatio-temporal information that enables association in time and space rather than purely using appearance information. 
Surprisingly our proposed method achieves better performance over several methods that leverage this additional spatio-temporal information, which further validates the reliability of our extracted features based on pose-aware multi-task learning. 

We also conducted an ablation study while comparing with state-of-the-art. 
It can be seen from the results that all the proposed algorithmic components, including multi-task learning and embedded pose representations, contribute to our performance gain. Though not all the components of our system contribute equally to the improved results, they all deliver viewpoint-aware information to aid feature learning. 
The combination of both triplet loss and cross-entropy loss outperforms the individual loss functions, because the metric in feature space and identity classification are jointly learned. 
The classification loss in ReID itself is generally too ``lazy'' to capture the useful but subtle attribute cues besides global appearance. 
Moreover, we experimented with DenseNet201, which has almost twice as many parameters compared to DenseNet121, but the results did not improve and even decreased due to overfitting. 
Therefore, the importance of the specific structure of HRNet for pose estimation is validated.
Finally,  we find that the additional synthetic data can significantly improve the ReID performance.

\begin{figure}[t]
\begin{center}
\includegraphics[width=1.0\linewidth]{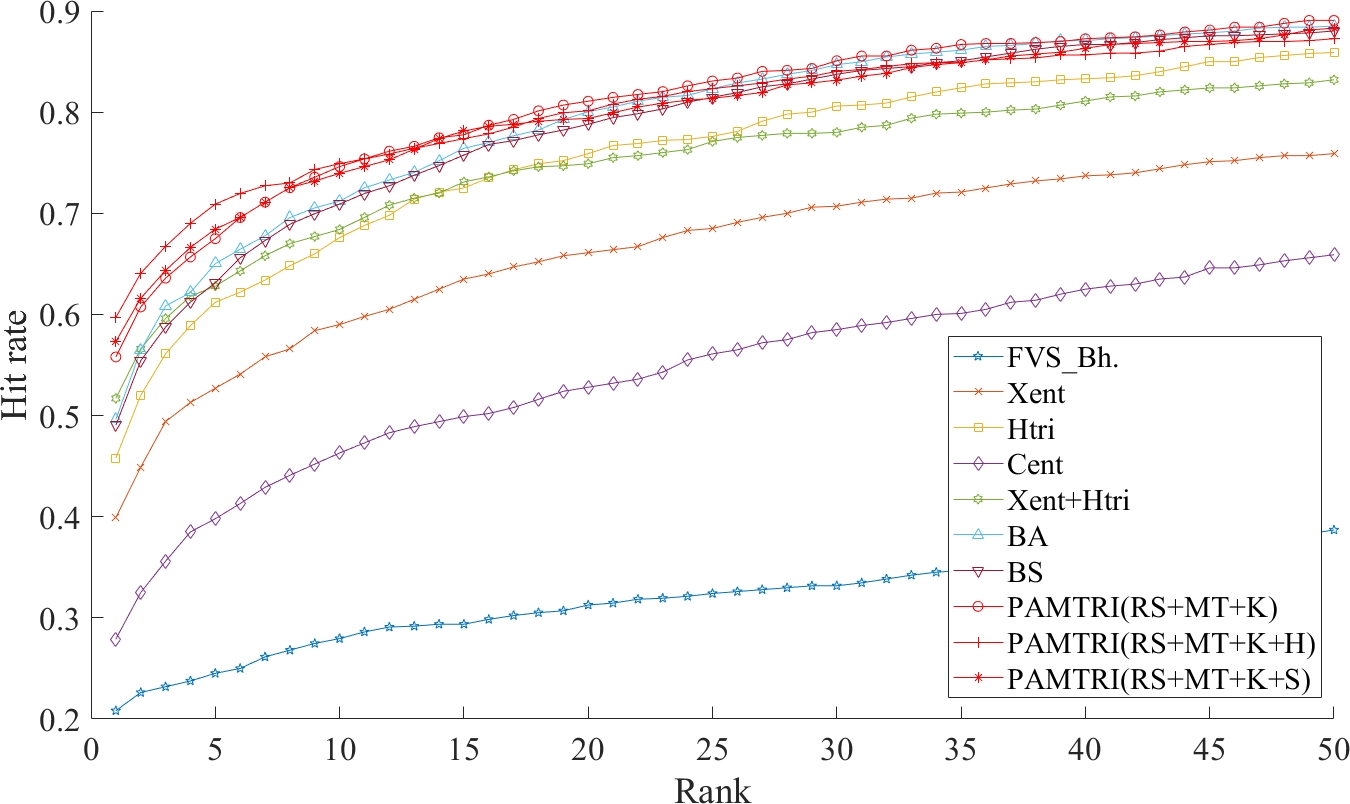}
\end{center}
   \caption{The CMC curves of state-of-the-art methods on CityFlow-ReID~\cite{Tang19CityFlow}.  Note that variants of our proposed method improve the state-of-the-art performance.  \textbf{Best viewed in color.}}
\label{fig:cmc}
\end{figure}

Tab.~\ref{tab:cityflow}, compares PAMTRI with  state-of-the-art on the CityFlow-ReID~\cite{Tang19CityFlow} benchmark. 
Notice the drop in accuracy of state-of-the-art compared to VeRi, which validates that this dataset is more challenging. 
BA and BS~\cite{Kumar19Vehicle}, which rely on triplet embeddings, are the same methods shown in the previous table for VeRi. 
Besides, we also compare with state-of-the-art metric learning methods in person ReID~\cite{Zhou18Torchreid} using cross-entropy loss (Xent)~\cite{Hermans17In}, hard triplet loss (Htri)~\cite{Szegedy16Rethinking}, center loss (Cent)~\cite{Wen16A}, and the combination of both cross-entropy loss and hard triplet loss (Xent+Htri). 
Like ours, they all share DenseNet121~\cite{Huang17Densely} as the backbone CNN. 
Finally, FVS~\cite{Tang18SingleCamera} is the winner of the vehicle ReID track at the AI City Challenge 2018~\cite{Naphade18The}. This method directly extracts features from a pre-trained network and computes their distance with Bhattacharyya norm. 

As shown in the experimental results, PAMTRI significantly improves upon state-of-the-art performance by incorporating pose information with multi-task learning. 
Again, all the proposed algorithmic components contribute to the performance gain. 
The experimental results of other ablation study align with the trends in Tab.~\ref{tab:veri}. 

\begin{figure*}[t]
\begin{center}
\includegraphics[width=0.95\linewidth]{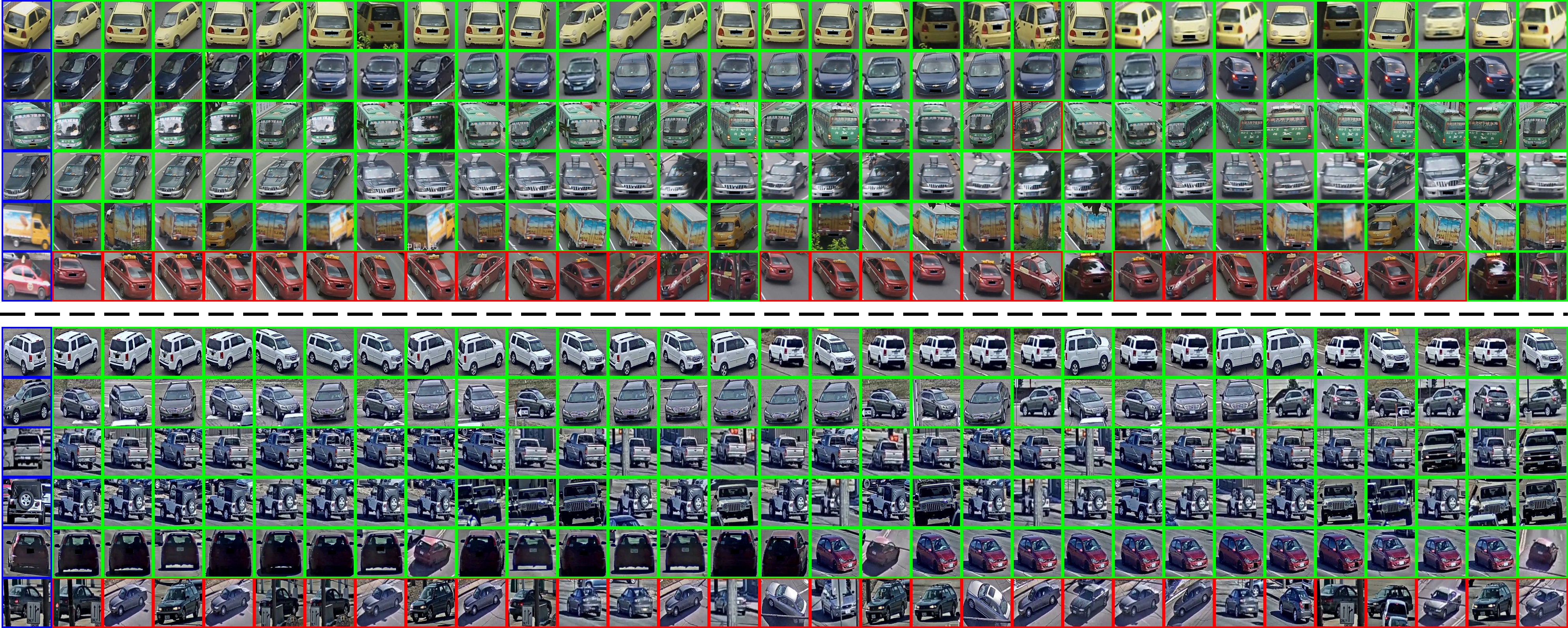}
\end{center}
   \caption{Qualitative visualization of PAMTRI's performance on public benchmarks:  VeRi (top rows, using RS+MT+K+S) and CityFlow-ReID (bottom rows, using RS+MT+K+H). For each dataset, 5 successful cases and 1 failure case are presented. For each row, the top 30 matched gallery images are shown for each query image (first column, blue). Green and red boxes represent the same identity (true) and different identities (false), respectively.  \textbf{Best viewed in color.}}
\label{fig:qual}
\end{figure*}

In Fig.~\ref{fig:cmc}, the CMC curves of the methods from Tab.~\ref{tab:cityflow} are plotted to better view the quantitative experimental comparison. 
We also show in Fig.~\ref{fig:qual} some examples of successful and failure cases using our proposed method. 
As shown in the examples, most failures are caused by high inter-class similarity for common vehicles like taxi and strong occlusion by objects in the scene, (\eg, signs and poles). 
\begin{table}
\begin{footnotesize}
\begin{center}
\begin{tabular}{lcccc}
\toprule
\multirow{2}*{Method} & \multicolumn{2}{c}{VeRi}  & \multicolumn{2}{c}{CityFlow-ReID} \\
\cmidrule{2-5}
 & Color acc. & Type acc. & Color acc. & Type acc.\\
\midrule
RS+MT & 93.42 & 93.27 & 80.16 & 78.97 \\
RS+MT+K & 93.86 & \textbf{93.53} & 83.06 & 79.17 \\
RS+MT+K+H & 94.06 & 92.77 & \textbf{84.80} & \textbf{80.04} \\
RS+MT+K+S & \textbf{94.66} & 92.80 & 83.47 & 79.41 \\
R+MT+K & 74.99 & 90.38 & 79.56 & 76.84 \\
\bottomrule
\end{tabular}
\end{center}
\caption{Experimental results of different variants of PAMTRI on color and type classification. The percentage of accuracy is shown. MT, K, H, S, RS and R respectively denote multi-task learning, explicit keypoints embedded, heatmaps embedded, segments embedded, training with both real and synthetic data, and training with real data only.}
\label{tab:attribute}
\end{footnotesize}
\end{table}

\subsection{Comparison of attribute classification}

Experimental results of color and type classification are given in Tab.~\ref{tab:attribute}. 
The evaluation metric is the accuracy in correctly identifying attributes. 
Again, these results confirm that CityFlow-ReID exhibits higher difficulty compared to VeRi, due to the diversity of viewpoints and environments. 
We also observe that the accuracy of type prediction is usually lower than that of color prediction, because some vehicle types look similar from the same viewpoint, \eg, a hatchback and a sedan from the same car model are likely to look the same from the front. 

It is worth noting that the pose embeddings significantly improve the classification performance. 
As explained in Sec.~\ref{sec:mtlearning}, pose information is directly linked with the definition of vehicle type, and the shape deformation by segments enables color estimation on the main body only.

In general, the accuracy of attribute classification is much higher than that of identity recovery, which could be used to filter out vehicle pairs with low matching likelihood, and thus improve the computational efficiency of target association across cameras. 
We leave this as future work.

\subsection{Comparison of vehicle pose estimation}

To evaluate vehicle pose estimation in 2D, we follow similar evaluation protocol as human pose estimation~\cite{Andriluka142D}, in which the threshold of errors is adaptively determined by the object's size. 
The standard in human-based evaluation is to use 50\% of the head length which corresponds to 60\% of the diagonal length of the ground-truth head bounding box.
Unlike humans, all the lengths between vehicle keypoints may change abruptly corresponding to viewing perspectives. 
Therefore, we use 25\% of the diagonal length of the entire vehicle bounding box as the reference, whereas the threshold is set the same as human-based evaluation. 
For convenience, we divide the 36 keypoints into 6 body parts for individual accuracy measurements, and the mean accuracy of all the estimated keypoints is also presented. 

\begin{table*}
\begin{footnotesize}
\begin{center}
\begin{tabular}{ccccccccc}
\toprule
Test set & Train set & Wheel acc. & Fender acc. & Rear acc. & Front acc. & Rear win. acc. & Front win. acc. & Mean\\
\midrule
\multirow{4}*{VeRi} & VeRi & \textbf{85.10} & 81.14 & 69.20 & 77.44 & 85.67 & \textbf{89.92} & 82.15 \\
  &  CityFlow & 58.62 & 54.99 & 45.32 & 54.86 & 65.74 & 74.38 & 60.14 \\
  & VeRi+Synthetic & 84.93 & \textbf{82.66} & \textbf{71.73} & \textbf{77.72} & \textbf{86.41} & 89.86 & \textbf{83.16} \\
  & CityFlow+Synthetic & 64.03 & 59.73 & 45.10 & 54.73 & 63.93 & 76.14 & 62.13 \\
\midrule
\multirow{4}*{CityFlow} & VeRi & 70.89 & 60.68 & 46.66 & 48.34 & 56.77 & 63.51 & 58.27 \\
  &  CityFlow & 83.75 & 79.89 & 65.87 & 71.48 & 75.38 & 80.80 & 77.07 \\
  & VeRi+Synthetic & 69.77 & 61.68 & 52.40 & 52.07 & 63.00 & 65.92 & 61.03 \\
  & CityFlow+Synthetic & \textbf{84.19} & \textbf{80.91} & \textbf{70.18} & \textbf{72.37} & \textbf{78.35} & \textbf{82.12} & \textbf{78.70} \\
\bottomrule
\end{tabular}
\end{center}
\caption{Experimental results of pose estimation using HRNet~\cite{Sun19Deep} as backbone network. The 36 keypoints are grouped into 6 categories for individual evaluation. Shown are the percentage of keypoints located within the threshold; see text for details.}
\label{tab:pose}
\end{footnotesize}
\end{table*}

\begin{figure*}[t]
\begin{center}
\includegraphics[width=1.0\linewidth]{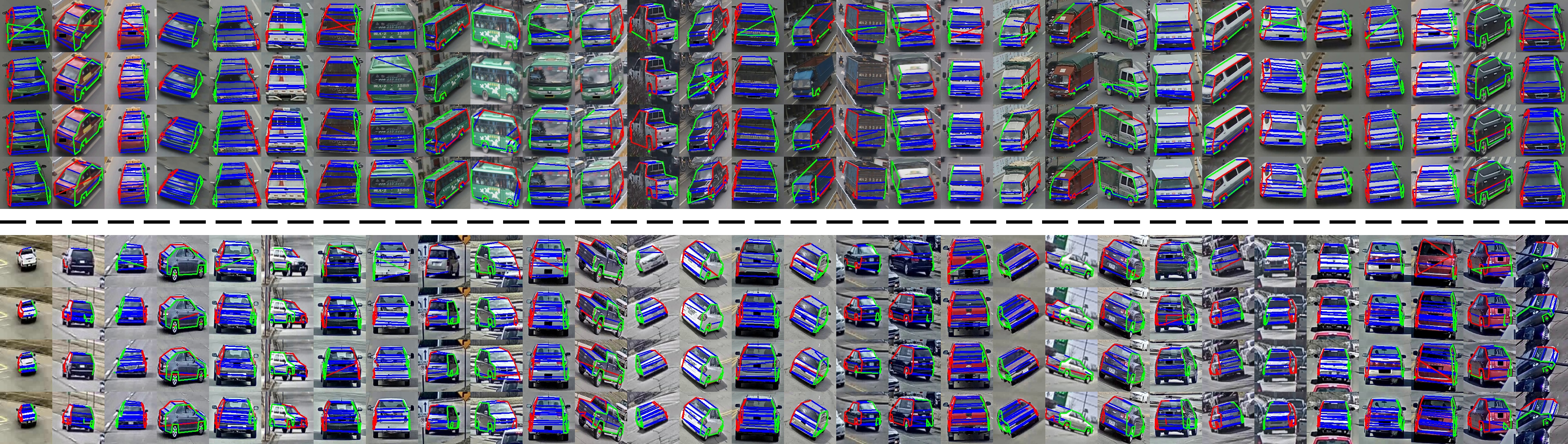}
\end{center}
   \caption{Qualitative visualization of the performance of pose estimation (only high-confidence keypoints shown). The top 4 rows show results on VeRi, whereas the bottom 4 show results on CityFlow-ReID.  For each, the rows represent the output from a different training set: VeRi, CityFlow-ReID, VeRi+synthetic, and CityFlow-ReID+synthetic, respectively. \textbf{Best viewed in color.}}
\label{fig:pose}
\end{figure*}

We randomly withhold 10\% of the real annotated identities to form the test set. 
The training set consists of synthetic data and the remaining real data.
Our experimental results are displayed in Tab.~\ref{tab:pose}. 
It is important to note that though the domain gap still exists in pose estimation, the combination with synthetic data can help mitigate the inconsistency across real datasets. 
In all the scenarios compared, when the network trained on one dataset is tested on the other, the keypoint accuracy increases as synthetic data are added during training. 
On the other hand, when the network model is trained and tested on the same dataset, the performance gain is more obvious on CityFlow-ReID, because the synthetic data look visually similar. 
Even with VeRi, improved precision can be seen in most of the individual parts, as well as the mean. 

From these results, we learn that the keypoints around the wheels, fenders, and windshield areas are more easily located, because of the strong edges around them. 
On the contrary, the frontal and rear boundaries are harder to predict, as they usually vary across different car models. 

Some qualitative results are demonstrated in Fig.~\ref{fig:pose}.
Most failure cases are from cross-domain learning, and it is noticeable that incorporating synthetic data improves robustness against unseen vehicle models and environments in the training set. 
Moreover, as randomized lighting and occlusion are enforced in the generation of our synthetic data, they also lead to more reliable performance against such noise in the real world. 
\section{Conclusion}

In this work, we propose a pose-aware multi-task learning network called PAMTRI for joint vehicle ReID and attribute classification. 
Previous works either focus on one aspect or exploit metric learning and spatio-temporal information to match vehicle identities. 
However, we note that vehicle attributes such as color and type are highly related to the deformable vehicle shape expressed through pose representations. 
Therefore, in our designed framework, estimated heatmaps or segments are embedded with input batch images for training, and the predicted keypoint coordinates and confidence are concatenated with the deep learning features for multi-task learning. 
This proposal relies on heavily annotated vehicle information on large-scale datasets, which has not yet been available. 
Hence, we also generate a highly randomized synthetic dataset, in which a large variety of viewing angles and random noise such as strong shadow, occlusion, and cropped images are simulated. 
Finally, extensive experiments are conducted on VeRi~\cite{Liu16A} and CityFlow-ReID~\cite{Tang19CityFlow} to evaluate PAMTRI against state-of-the-art in vehicle ReID. 
Our proposed framework achieves the top performance in both benchmarks, and an ablation study shows that each proposed component helps enhance robustness. 
Furthermore, experiments show that our schemes also benefit the sub-tasks on attribute classification and vehicle pose estimation. 
In the future, we plan to study how to more effectively bridge the domain gap between real and synthetic data.

{\small
\bibliographystyle{ieee_fullname}
\bibliography{main}
}

\end{document}